\documentclass{article}

\usepackage[nonatbib]{neurips_2023}

\usepackage[utf8]{inputenc} % allow utf-8 input
\usepackage[T1]{fontenc}    % use 8-bit T1 fonts
\usepackage{hyperref}       % hyperlinks
\usepackage{url}            % simple URL typesetting
\usepackage{booktabs}       % professional-quality tables
\usepackage{amsfonts}       % blackboard math symbols
\usepackage{nicefrac}       % compact symbols for 1/2, etc.
\usepackage{microtype}      % microtypography
\usepackage{xcolor}         % colors

\usepackage{graphicx}       %pictures

\title{On enhancing the robustness of Vision Transformers: Defensive Diffusion}

\author{
Raza Imam, Muhammad Huzaifa, Mohammed El-Amine Azz \\
MBZUAI\\
\texttt{[raza.imam, muhammad.huzaifa, mohammed.azz]}\texttt{@mbzuai.ac.ae}
}

\begin{document}
% * \footnote{*: Authors with equal contribution}

\maketitle

\begin{abstract}
    Privacy and confidentiality of medical data are of utmost importance in healthcare settings. ViTs, the SOTA vision model, rely on large amounts of patient data for training, which raises concerns about data security and the potential for unauthorized access. Adversaries may exploit vulnerabilities in ViTs to extract sensitive patient information and compromising patient privacy. This work address these vulnerabilities to ensure the trustworthiness and reliability of ViTs in medical applications. In this work, we introduced a defensive diffusion technique as an adversarial purifier to eliminate adversarial noise introduced by attackers in the original image. By utilizing the denoising capabilities of the diffusion model, we employ a reverse diffusion process to effectively eliminate the adversarial noise from the attack sample, resulting in a cleaner image that is then fed into the ViT blocks. Our findings demonstrate the effectiveness of the diffusion model in eliminating attack-agnostic adversarial noise from images. Additionally, we propose combining knowledge distillation with our framework to obtain a lightweight student model that is both computationally efficient and robust against gray box attacks. Comparison of our method with a SOTA baseline method, SEViT, shows that our work is able to outperform the baseline. Extensive experiments conducted on a publicly available Tuberculosis X-ray dataset validate the computational efficiency and improved robustness achieved by our proposed architecture. Our code is publicly available at \url{https://github.com/Muhammad-Huzaifaa/Defensive_Diffusion}.

\end{abstract}

\section{Introduction}

The increasing use of deep learning in medical imaging has brought attention to the vulnerabilities present in CNN and ViT models. The susceptibility of ViTs to adversarial, privacy, and confidentiality attacks raises significant concerns regarding their dependability in medical applications. The use of AI in healthcare introduces vulnerabilities in medical imaging systems, as malicious actors may exploit adversarial attacks for financial gain through fraudulent billing or insurance claims. This necessitates the development of robust defensive strategies to ensure the secure deployment of automated medical imaging systems. Adversarial attacks pose a security threat where attackers manipulate input data to deceive machine learning models, leading to incorrect predictions \cite{goodfellow2014explaining}. Detecting and defending against these attacks is crucial for accurate diagnoses and treatment outcomes.

Vision Transformers (ViTs)\cite{vaswani2017attention} have become one of the best-performing models in computer vision tasks such as segmentation and classification, performing better in some instances than traditional CNNs. CNNs' vulnerability to adversarial attacks is a well-known problem, however, in terms of robustness, \cite{bhojanapalli2021understanding} research has shown that ViTs are also vulnerable to different types of adversarial attacks that can change the classification output with minimal change to the appearance of the images to the human eye. This issue can be detrimental in essential applications such as medical imaging, insurance fraud detection, and self-driving cars. As such, presenting the ensembling ViT variant (SEVIT) can have positive outputs for robustness and accuracy for the models.  However, SEViTs require a lot of computational power compared to their original ViT. So what if we utilize defensive measures against adversarial attacks to improve ViTs to lower the computational requirements for the performance of SEViTs. We aim to explore the robustness of ViTs and SEViTs under different types of attacks, as well as to utilize defensive measures, such as diffusion models and knowledge distillation to further improve the robustness and transferability of the models. This can help us better understand how ViTs and SEViTs work, as well as study the trade-off required for the use of SEViTs. Furthermore, studying the defensive measures can help us improve model architecture to build improved models.

The privacy and confidentiality of medical data are crucial in healthcare settings. Vision Transformers (ViTs), being state-of-the-art vision models, raise concerns about data security and unauthorized access due to their reliance on large patient datasets for training. Attackers can exploit vulnerabilities in ViTs to extract sensitive patient information, jeopardizing privacy. SEVIT \cite{almalik2022self}, an ensemble-based variant of Vision Transformers, have demonstrated promising robustness in medical image analysis compared to traditional ViTs. However, SEVITs still have vulnerabilities that compromise their accuracy and reliability, particularly in terms of privacy and confidentiality. Attacks like Model Extraction pose a significant threat to patient privacy by allowing adversaries to extract SEVIT models and misuse them for various purposes, such as selling them or inferring sensitive patient information from X-ray images. Despite their effectiveness against adversarial attacks, SEVITs overlook the aspect of Trustworthy AI, making them insufficient as a comprehensive solution. While SEVIT has shown robust performance against adversarial attacks, its large MLP modules pose efficiency challenges for real-world deployment. This research signifies on enhancing the robustness of the SEVIT model for Tuberculosis X-ray classification by leveraging defensive diffusion and adversarial training techniques. The objective of this work is to improve both computational efficiency and the overall robustness of medical imaging systems using vision transformers. Specifically, we introduce a defensive diffusion technique as an adversarial purifier to remove adversarial noise introduced by attackers in the original image. By leveraging the denoising capabilities of the diffusion model, we apply a reverse diffusion process to effectively eliminate adversarial noise, resulting in a cleaner image that is then fed into the ViT blocks. We aim to demonstrate the efficacy of the diffusion model in eliminating attack-agnostic adversarial noise from images. Combining knowledge distillation with our framework to obtain a lightweight student model that is both computationally efficient and robust against gray box attacks \cite{carlini2017towards}. Achieving benchmark-level robustness is crucial for enhancing diagnosis and treatment outcomes, benefiting both patients and healthcare providers. Our comparative analysis with a state-of-the-art baseline method, SEViT, aims to validate our method in terms of efficiency and robustness. The structure of this work is organized as follows: Section 2 provides a review of relevant literature. Section 3 describes the data exploration and preprocessing stages. Section 4 introduces the proposed method. Section 5 presents the comprehensive experiments conducted and the analyses performed. The results and findings of these experiments are discussed in Section 6. Finally, Section 7 concludes the work and proposes potential avenues for future research.

\begin{figure}[htpb]
    \centering
    \includegraphics[width=12cm]{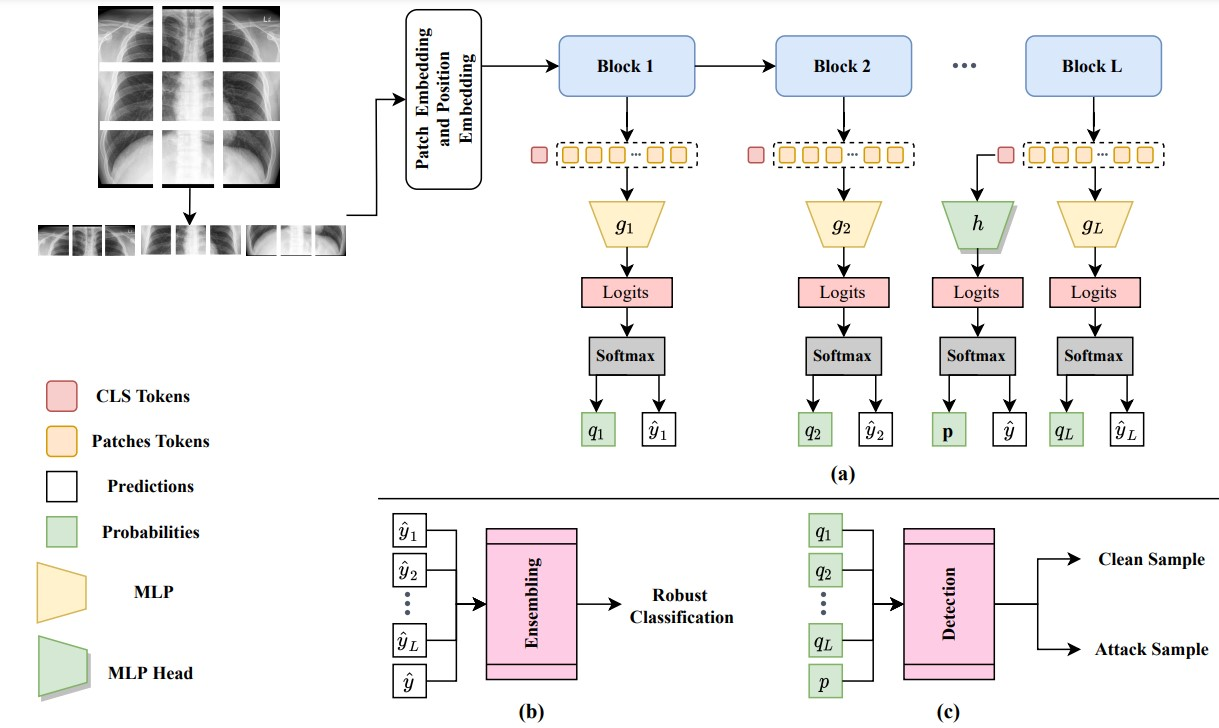}
    \caption{The baseline SEViT architecture involves extracting patch tokens from the initial blocks and training separate MLPs as illustrated in (a). A self-ensemble of MLPs is utilized in combination with the final ViT classifier, as depicted in (b). Consistency among the predictions made by the ensemble can be leveraged to detect adversarial samples \cite{almalik2022self}.}
    \label{fig:sevit}
\end{figure}

\section{Literature Review}

\subsection{SEViTs}
Self-Ensembling Vision Transformer (SEVIT) is a recent development in computer vision that combines the benefits of both Vision Transformers (ViT) and Self-Ensembling models. SEVIT utilizes the ViT architecture for image feature extraction and employs a self-ensembling technique to improve the model's generalization and performance Figure \ref{fig:sevit}. \cite{almalik2022self}, The Self-Ensembling Vision Transformer (SEViT) proposed by the author makes use of the observation that the feature representations learned by the early blocks of a ViT are not easily impacted by adversarial perturbations. The SEViT approach involves training multiple classifiers based on these intermediate feature representations and combining their predictions with that of the final ViT classifier to improve resistance against adversarial attacks. Additionally, the consistency between the various predictions can be used to identify potential adversarial samples. Through experiments on chest X-ray and fundoscopy modalities, the effectiveness of the SEViT architecture is demonstrated in defending against various adversarial attacks in the gray-box scenario, where the attacker has knowledge of the target model but not the defense mechanism.

\subsection{Adverserial Attacks}
Adversarial attacks are a class of techniques that aim to fool machine learning models by adding imperceptible perturbations to the input data. Two popular adversarial attacks are the fast gradient sign method (FGSM) and the projected gradient descent (PGD) attack. The FGSM attack, proposed by \cite{goodfellow2014explaining}, generates an adversarial example by adding a small perturbation to the input data in the direction of the sign of the gradient of the loss function with respect to the input. Despite its simplicity, the FGSM attack has been shown to be effective in generating adversarial examples for various image classification tasks. The PGD attack, proposed by \cite{madry2017towards}, is an iterative version of the FGSM attack that generates adversarial examples by taking multiple small steps in the direction of the sign of the gradient of the loss function while constraining the perturbation to be within a certain bound. The PGD attack has been shown to be more effective than FGSM in generating adversarial examples that are harder for the model to classify correctly, even when the model has been trained to defend against FGSM attacks. Studies have explored different aspects of these attacks, such as their transferability across different models and datasets, their effectiveness against models trained with different defenses, and their impact on the interpretability and robustness of machine-learning models \cite{papernot2016limitations} \cite{carlini2017towards}

\subsection{Diffusion Models}
Model diffusion is a technique that has been gaining increasing attention in the field of computer vision. It involves the propagation of information between different images, allowing for the transfer of style, texture, and other features from one image to another. One of the earliest papers on model diffusion was \cite{matsunaga2022fine} in which the authors propose a technique for fine-grained image editing using diffusion models. Their approach involves propagating the information from a reference image to a target image in a pixel-wise manner, allowing for precise control over the editing process. The authors demonstrate that their technique can be used for tasks such as image inpainting, super-resolution, and style transfer. In \cite{nie2022diffusion}, the authors propose a method for removing adversarial perturbations from images using diffusion models. Their approach involves first adding the Gaussian noise to adversarially perturb images and then using the denoising diffusion model to reconstruct back the image. The authors demonstrate that their method is effective in improving the robustness of deep neural networks against adversarial attacks. Other papers have also explored Face Recognition \cite{ren2022learning} and image segmentation. In \cite{bansal2022cold} the authors explored the diffusion model by adding a different variation of noises including blurring, snowing, and masking, and showed that the diffusion model is still robust enough to reconstruct back the original image. Their method called 'cold diffusion' is based on the heat diffusion equation and the concept of coldness or lack of randomness. The authors demonstrated the effectiveness of their method through experiments on several datasets, including MNIST and CIFAR-10. In \cite{chambon2022adapting}, the authors proposed a method for adapting pre-trained vision-language models to medical imaging domains. They leveraged the transformer-based model's ability to encode global contextual information and applied it to medical images by formulating the task as a multi-label classification problem. The authors demonstrated the effectiveness of their approach on several medical imaging datasets, achieving state-of-the-art results on some of them. This work highlights the potential of diffusion models in adapting pre-trained models to new domains, enabling the use of existing models for new applications.

\subsection{Knowledge Distillation}

Knowledge distillation is a process of training a smaller, faster, and more efficient model, known as the student model, to mimic the behavior of a larger and more accurate model, known as the teacher model. This technique has gained significant attention in recent years due to its ability to compress complex models, reduce their computational requirements, and accelerate their inference time. One of the earliest works on knowledge distillation was proposed by \cite{hinton2015distilling}, where they introduced the concept of transferring the knowledge of a deep neural network to a smaller network by minimizing the difference between the outputs of the teacher and student models. In their work, they demonstrated that knowledge distillation could significantly improve the performance of a smaller network on various image classification tasks. Following this work, several studies have explored different aspects of knowledge distillation, such as the choice of the loss function, the impact of the teacher and student model architectures, and the effect of data augmentation techniques. One such work is the study by \cite{zagoruyko2016paying}, where they proposed a novel loss function, called the attention transfer loss, which leverages the attention maps of the teacher model to guide the training of the student model. Their results showed that attention transfer loss could significantly improve the performance of the student model on various image classification tasks. Moreover, recent research has extended knowledge distillation to other domains, such as natural language processing, speech recognition, and object detection \cite{sanh2019distilbert}.

\section{Data exploration and preprocessing}
To evaluate the robustness of ViTs and SEViTs against defensive measures, we employed the Tuberculosis (TB) Chest X-ray Database \cite{rahman2020reliable}. This database consists of high-resolution images and contains around 6500 images with two classes: Normal and Tuberculosis. The dataset was complete without any missing data, eliminating the need for data cleaning. The images were originally in the format of 512 by 512 pixels, with three color channels. Prior to analysis, we preprocessed the images to a size of 224 by 224 pixels. The dataset was divided into three parts: training, validation, and testing, with approximately 5000, 700, and 700 images, respectively. The split was performed randomly, allocating 80\% of the images for training, 10\% for validation, and the remaining 10\% for testing. This random division of the dataset ensures that the training, validation, and testing sets are mutually exclusive, allowing for the evaluation of the model's ability to generalize to new, unseen images.

\section{Methodology}
In this section, we provide the detailed overview of our proposed method, defensive diffusion, to defend against adversarial attacks, along with a lightweight model obtained through knowledge distillation, which reduces computational complexity without sacrificing performance.
\subsection{Hypothesis}
We hypothesis that (1) reverse diffusion process is effective in removing adversarial noise, (2) adversarial images generated specifically for the vision transformer would not be transferable to the student model (CNN). Faris et al. \cite{almalik2022self} conducted a study revealing that Vision Transformers (ViTs) lack robustness against white box attacks. In our research, we address this issue from a different perspective by leveraging the capabilities of a diffusion model to mitigate attack-agnostic noise. By exploiting the denoising properties of the diffusion model, we employ a reverse diffusion process to remove adversarial noise and obtain a clean image.

In a gray box scenario, where the attacker possesses comprehensive knowledge of the base model, including model weights, prediction probabilities, and training data, but lacks information about the defensive model, \cite{almalik2022self} propose the SEViT model. SEViT incorporates trained Multi layer Perceptions (MLPs) attached to each block of the vision transformer to enhance robustness. However, deploying this approach becomes infeasible in real-world due to computational and memory complexities, as each MLP block contains a substantial number of parameters (625M). To overcome this challenge, our methodology incorporates performing knowledge distillation to adopt a lightweight student CNN model trained on soft probabilities distilled from the teacher model (ViT). Our hypothesis aligns with the observations made by \cite{naseer2021intriguing}, who demonstrated that adversarial images exhibit higher transferability within the same family of model variants.

\subsection{Methods}
\subsubsection{Defensive Diffusion}
Recent studies \cite{dhariwal2021diffusion} \cite{rombach2022highresolution} have demonstrated that diffusion models surpass Generative Adversarial Networks (GANs) in terms of image generation and image quality. The research focus has shifted towards generative diffusion models due to their ability to generate images from noise. Prior works have explored the potential of diffusion models as purifiers by utilizing their reverse diffusion process to eliminate noise from images \cite{nie2022diffusion}. Inspired by these advancements \cite{xiao2022densepure}, we propose a novel approach called 'Defensive Diffusion' that effectively eliminates adversarial noise from images and restores model accuracy. The diffusion model comprises both forward and reverse diffusion processes. In the forward diffusion process, the model progressively introduces small amounts of Gaussian noise to the image at each time step, following a specific equation \ref{eqn:forward}, until the image is completely transformed into noise \cite{sohl2015deep}:. 
\begin{equation}
\label{eqn:forward}
q(x_t|x_{t-1}) = N(x_t;\sqrt{1-B_t}x_{t-1},B_tI)
\end{equation}
where q is the forward diffusion image at time step conditioned on the previous time step,$B_t$ is the forward process variance that can be learned via reparameterization and $\mu_\theta$ is the mean function approximator \cite{ho2020denoising}.

The diffusion model is then trained to reconstruct the original image from the noise in the reverse diffusion process, leveraging knowledge from the training distribution. The equation governing the reverse diffusion process is as follows:
\begin{equation}
\label{eqn:reverse}
p(x_{t-1}|x_t) = N(x_{t-1};\mu_{\theta}(x_t,t),\sum_{\theta}(x_t,t))
\end{equation}

\begin{figure}[htbp]
    \centering
    \includegraphics[width=14cm]{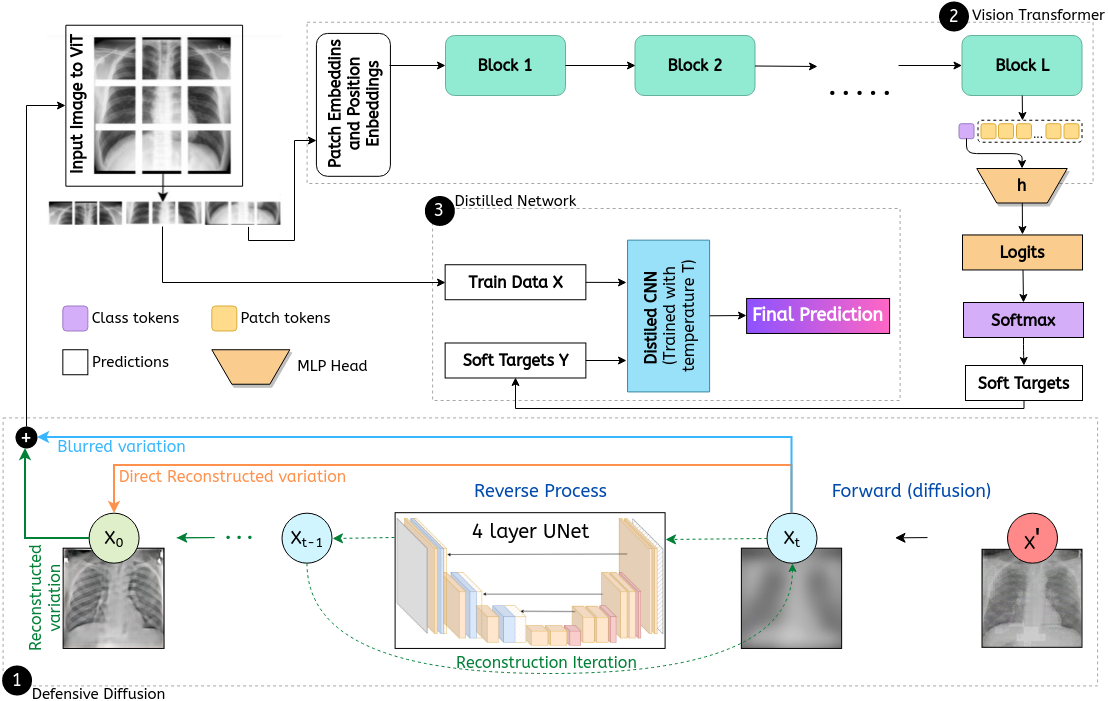}
    \caption{The proposed framework of Defensive Diffusion for enhanced ViT robustness. (1) Diffusion purifier block that utilizes the forward-reverse phases to generate three variations of diffusion. (2) Purified image generated from step 1 is inputted to Vision Transformer block where tokens are extracted for training to eventually obtain the soft targets. (3) Using the soft predictions attained in step 2, knowledge distillation is performed on a new student (4-layer CNN) model to obtain the final predictions.}
    \label{fig:method}
\end{figure}

In our approach, we begin by training the deblurring diffusion model in an end-to-end manner \cite{bansal2022cold}, using the training set from the TB-dataset. We then apply this trained model to process adversarial images generated using a vision transformer. The forward diffusion process is employed to introduce blur to the images, effectively mixing the adversarial noise with the blurred noise. The resulting blurred images are subsequently fed into the deblurring diffusion model to obtain denoised images as shown in Figure \ref{fig:method}. To evaluate the effectiveness of our method, we compare the quality of the blurred and reconstructed images. Unlike the approach proposed in \cite{nie2022diffusion}, which focuses on adversarial purification, our method leverages the deblurring diffusion model. Here we consider three variation of diffusion model phases. In the 'blurred phase' we evaluate the final output obtained from forward diffusion process that is $x_{t}$ given by 

\begin{equation}
\label{eqn:for_blurr}
x_{t} = G_{t} \ast x_{t-1} = G_{t} \ast \ldots \ast G_{1} \ast x_{0} = \bar{G}_{t} \ast x_{0} = D(x_{0}, t),
\end{equation}
where * denotes the convolution operator, which blurs an image using a kernel. Deblurring can be thought of as adding frequencies to the image. 

In the 'algorithmic reconstruction' phase, we employ the method described in \cite{bansal2022cold}. This approach involves reconstructing the image by combining it with the image from the previous time step, followed by the addition of variable noise to the image until time t. As the time steps progress, the amount of added noise gradually decreases. To facilitate 'direct reconstruction', we leverage the trained U-net model, which allows us to reconstruct the clean samples in a single step. The equation for direct reconstruction is given by
\begin{equation}
\label{eqn:direct_rec}
R(D(x_0, T)
\end{equation}

where $x_0$ is the clean sample, D and R are the degradation and restoration operator. 

\subsubsection{Knowledge Distillation}
\begin{figure}[htbp]
    \centering
    \includegraphics[width=8cm]{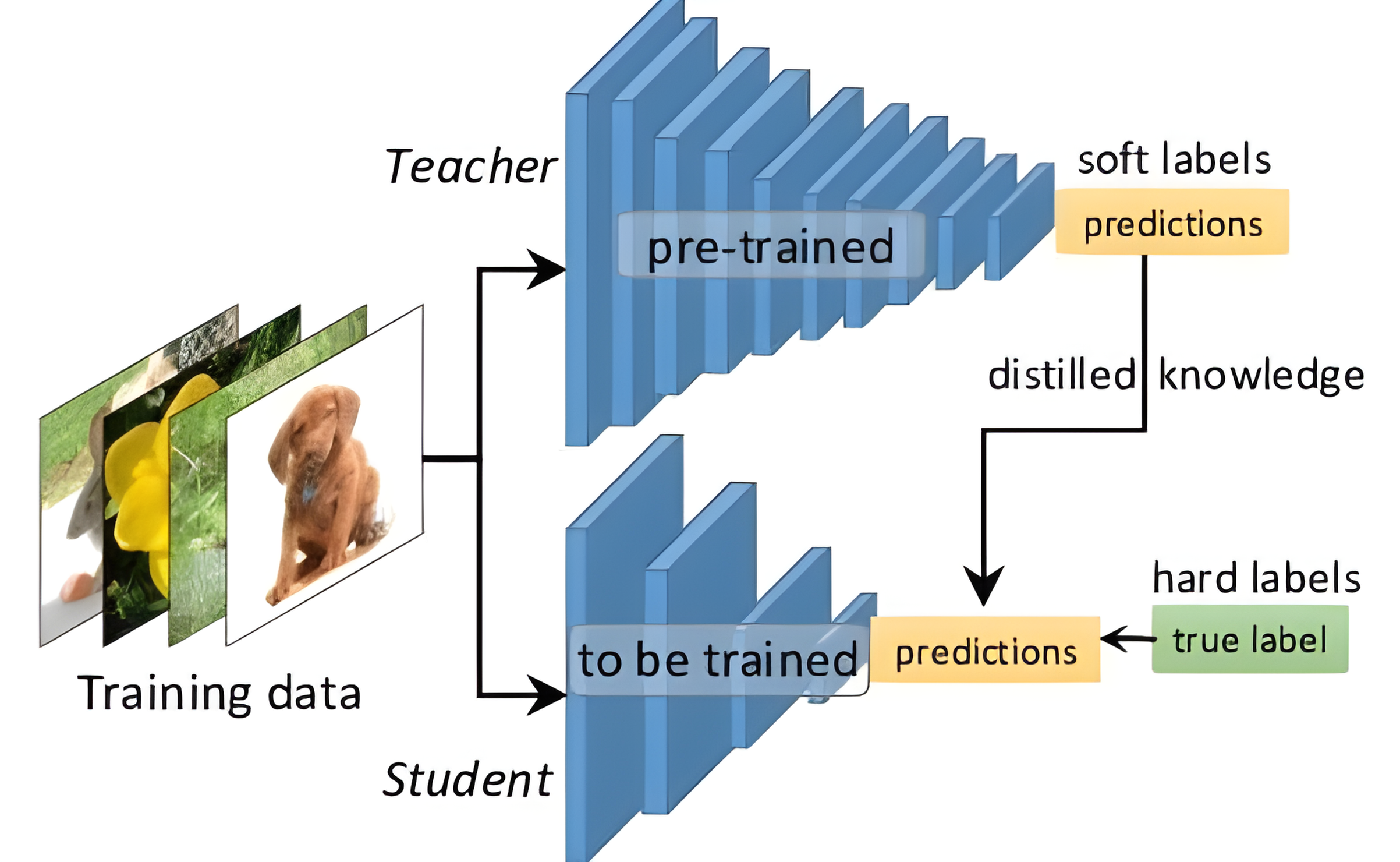}
    \caption{Knowledge distillation: Involves training a teacher model to generate a  smaller student model trained on soft labels.}
    \label{fig:knowledge distillation}
\end{figure}

Knowledge distillation involves training a teacher model to generate soft targets, which are then used to train a smaller student model\cite{hinton2015distilling}. The student model, also known as the distilled model, exhibits improved robustness due to the introduction of uncertainties in its output. This increased uncertainty makes it more challenging for attackers to generate adversarial examples that can deceive the model. The distilled model is trained using soft probabilities/targets instead of hard targets. Soft targets provide valuable information that cannot be conveyed through a single hard target. They effectively communicate the patterns and regularities discovered by a model trained on the entire dataset to another model, enabling better generalization. We use CNN because one of the properties of adversarial attack is that they are highly transferable with the family of models\cite{naseer2021intriguing}. \cite{hinton2015distilling} have demonstrated that training the model with hard probabilities lead to severe overfitting. Therefore, soft probabilities is used to obtain a more robust student model. Moreover, soft probabilities introduces a smoothness to the optimization landscape during training. The light weight CNN is trained on soft probabilities, input from the last layer of ViT as shown by the Figure \ref{fig:knowledge distillation}.

\section{Experimentations}
\textbf{Setup.} For employing the diffusion model, we incorporated deblurring stable diffusion by \cite{bansal2022cold} (2022) and tested it on ViT, SEViT, and distilled CNN model. The Vision Transformer (ViT) model we utilized is based on the ViT-B/16 architecture pretrained on the ImageNet dataset [5]. To implement SEViT and enable a baseline comparison, we trained 12 MLP modules, one for each ViT block, to serve as intermediate classifiers taking patch tokens as input. We used the fine-tuned ViT from the original SEViT method. For the distilled network, we designed a 3-layer CNN surrogate model trained on soft probabilities generated by the original ViT model. Our experiments were performed on a single Nvidia Quadro RTX 6000 GPU with 24 GB memory. The resulting ViT model achieved an accuracy of 96.3\% on the original clean test set of chest X-ray images.

\textbf{Attack types.} In this work, we have experimented with 3 adversarial attack algorithms. We employ the Foolbox library \cite{rauber2018foolbox} to generate these distinct types of $L_{\infty}$ adversarial attacks, namely FGSM \cite{goodfellow2014explaining}, PGD \cite{madry2017towards}, and AutoPGD \cite{croce2020reliable}. These attacks are created with a perturbation value of ${\epsilon}$=0.03, while maintaining default values for all other parameters, where this perturbation value implies the magnitude of 'adversarial-ness' introduced to the attack sample.
\textbf{}

\subsection{Defensive Diffusion}
\begin{figure}[htbp]
    \centering
    \includegraphics[width=12cm]{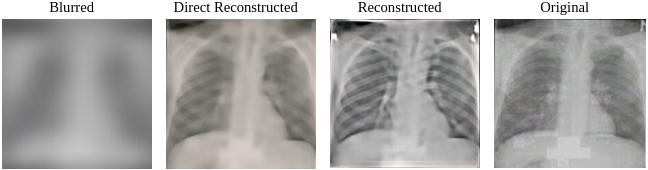}
    \caption{Types of images generated through Original + three diffusion variations via Diffusion phase}
    \label{fig:img_types}
\end{figure}

We implemented three variation of diffusion phases in diffusion model (where this diffusion phase consists of forward diffusion and reconstruction phases) on ViT. The three variation of diffusion phases are blurring, reconstruction and direct reconstruction. In the blurring variation, we take the image processed solely by the forward diffusion phase. In the reconstruction variation, we reconstruct the image from its blurred state back to a fine-grained image (See Figure \ref{fig:img_types}). Moreover, the direct reconstruction involves performing the reconstruction directly without any iterations. The diffusion model acts as a defensive purifier added before the ViT that aims to remove the adversarial noise in the input image. Our proposed diffusion framework, with its diffusion phase, aims to purify the input image before feeding it into the original ViT module for classification. We performed the same implementation of combining defensive diffusion phase with SEViT instead of ViT with the aim to compare with the baseline performance of robustness in both the models. Our experiments yield several noteworthy observations, as outlined in Table \ref{tab:diff_vit}. On the other hand, SEViT, which is originally built to be robust against attack samples, achieved significantly higher robust accuracy on attack samples 

\begin{table}[htbp]
\caption{Defensive diffusion when performed with ViT (left) and SEViT (right) across Clean and Attack (FGSM, PGD, AutoPGD) samples. Here Original are attack samples without any diffusion phase, whereas the other three variations (blurred, reconstructed and direct reconstructed) are samples generated when input attack samples are passed through the diffusion phase. In SEViT, we are ensembling 3 MLPs.}
\centering
\begin{minipage}{1\textwidth}
  \centering
  \includegraphics[width=9cm]{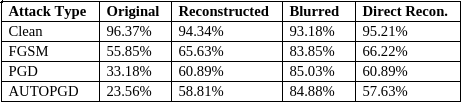}
\end{minipage}

\begin{minipage}{1\textwidth}
  \centering
  \includegraphics[width=9cm]{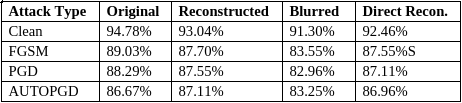}
\end{minipage}

\label{tab:diff_vit}
\end{table}

% \subsection{Diffusion on SEViT}
\subsection{Knowledge Distillation}
In order to improve robust accuracy while preserving clean accuracy of ViT, along with improved computational performance, we employed knowledge distillation to extract a student CNN model from the original teacher ViT. During the distillation process, we utilized the training set to train the student ViT, which generated soft targets for each input sample. These soft targets were then used to create a new dataset (X, Soft target) for training the distilled student CNN. A temperature parameter (set to 1 in our case) was employed to determine the degree of softness in the student model. We conducted the same defensive diffusion experiments on the student CNN as we did on ViT and SEViT. This involved implementing three diffusion variations as purifiers for the student CNN and subsequently testing it on both actual attack samples and diffused attack samples as shown in Table \ref{tab:vit_sevit_cnn}. 

\begin{table}[htbp]
    \caption{Clean and Robust performance on distilled student CNN}
    \centering
    \includegraphics[width=9cm]{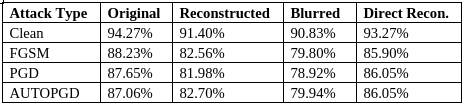}
    \label{tab:vit_sevit_cnn}
\end{table}

\section{Results and Discussions}
\subsection{Results on Diffusion with Raw ViT vs Diffusion ViT vs SEViT}
\begin{figure}[htbp]
\centering
\begin{minipage}{.5\textwidth}
  \centering
  \includegraphics[width=7cm]{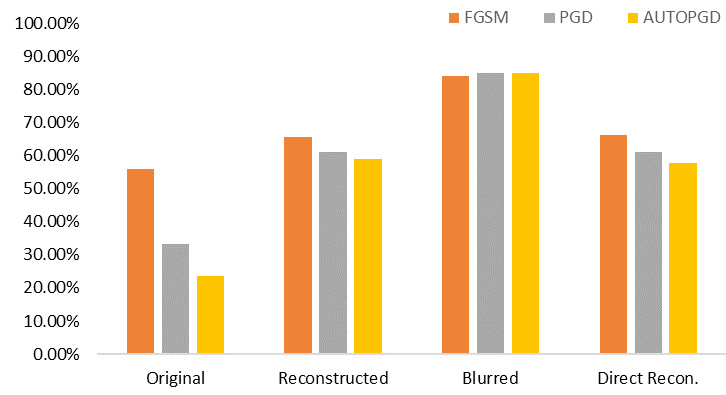}
\end{minipage}%
\begin{minipage}{.5\textwidth}
  \centering
  \includegraphics[width=7cm]{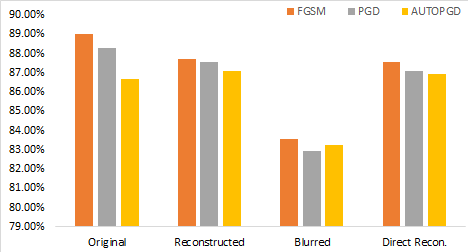}
\end{minipage}
\caption{Defensive diffusion when performed with ViT (left) and SEViT (right) across Attack samples. Here Original are attack samples without any diffusion phase, whereas the other three variations (blurred, reconstructed and direct reconstructed) are samples generated when input attack samples are passed through the diffusion phase. In SEViT, we are ensembling 3 MLPs. The attack samples are generate with ${\epsilon=0.03}$.}
\label{fig:diff_vit}
\end{figure}

We implemented our proposed diffusion method on both ViT and SEViT models, and the results are presented in Figure \ref{fig:diff_vit}. As a defender, our objective is to defend against attack samples while maintaining a robust accuracy close to the clean accuracy. This means that we want our model to accurately classify both clean and attack samples, ensuring correct predictions. In the case of original ViT, we observe that the inference using attack samples (Original) significantly decreases the clean accuracy to approximately 55\%. However, when utilizing our proposed diffusion purifier with ViT, we achieve higher robust accuracy across all three diffusion variations. By leveraging the denoising properties of the diffusion model, we apply a reverse diffusion process that effectively removes the adversarial noise from the attack sample, resulting in a cleaner image. Among the three diffusion variations, the blurred variation performs the best, achieving robust accuracy of approximately 85\% in the case of PGD attack. On the other hand, the reconstructed and direct reconstructed variations show lower robust accuracy, with around 61\% robust accuracy in the PGD attack scenario. Therefore, we imply that blurred variation of diffusion phase of our approach is able to produce enhanced robust accuracy when tested with attack samples compared to raw ViT (i.e, ViT without diffusion).

\begin{figure}[htbp]
    \centering
    \includegraphics[width=7cm]{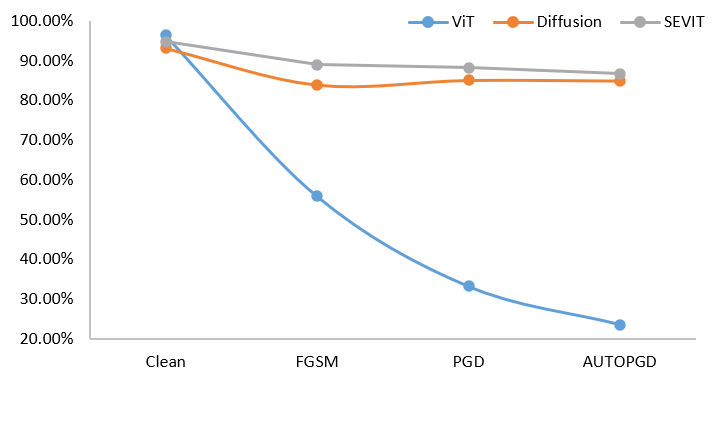}
    \caption{Robust performance of ViT on attacked images vs ViT with Defensive Diffusion (blurred images) vs SEViT on attacked images}
    \label{fig:vit_sevit_cnn}
\end{figure}

Furthermore, SEViT demonstrates higher robust accuracy even without the use of diffusion phases, with clean accuracy of approximately 94\% and attack accuracy of around 89\% in the case of FGSM attack. This higher robustness is consistent across other types of attacks as well. However, when the diffusion phase is applied to SEViT, there is a decrease in robust accuracy across all three diffusion variations. This reduction in robust accuracy can be attributed to the introduction of over-smoothing caused by the diffusion phases, which results in the loss of important details and features necessary for robust classification. This over-smoothing effect ultimately leads to reduced discriminative power and slightly lower robust accuracy when SEViT is combined with the diffusion phase. When considering the best performing variation, which is the blurred variation among the three diffusion phases, we observe that when the blurred diffusion variation is applied to ViT, it achieves a robust accuracy that is comparable to SEViT (See Figure \ref{fig:vit_sevit_cnn}). SEViT utilizes an ensemble of MLPs, with each MLP containing 625 million parameters. This introduces computational complexities and makes practical deployment infeasible. In contrast, our blurred variation of the diffusion phase demonstrates almost similar robust performance to SEViT, with only a 1-2\% lower difference in robust accuracy. This suggests that our approach of incorporating the blurred diffusion variation into ViT can be an efficient and robust solution for developing a ViT model that is both effective and practical. By using the blurred diffusion variation, we can achieve similar levels of robustness to SEViT without the computational complexity associated with the ensemble of MLPs.

\subsection{Results on Teacher ViT vs Student CNN}
\begin{figure}[htbp]
    \centering
    \includegraphics[width=7cm]{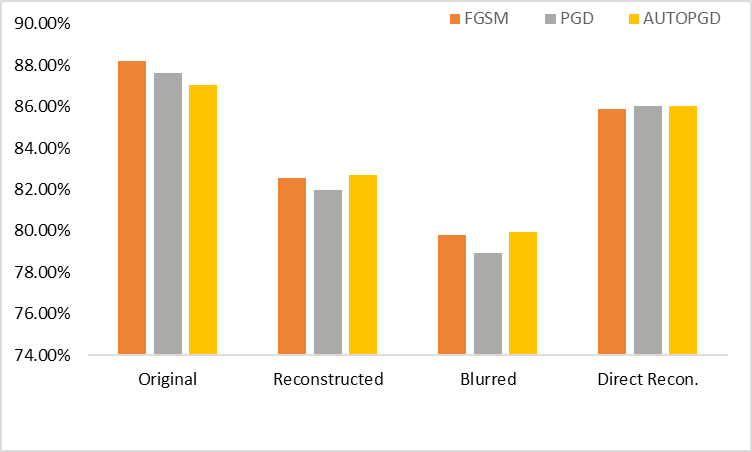}
    \caption{Different phases of defensive diffusion tested on distilled CNN compared with the clean images}
    \label{fig:diff_cnn}
\end{figure}

\begin{figure}[htbp]
\centering
\begin{minipage}{.5\textwidth}
  \centering
  \includegraphics[width=7cm]{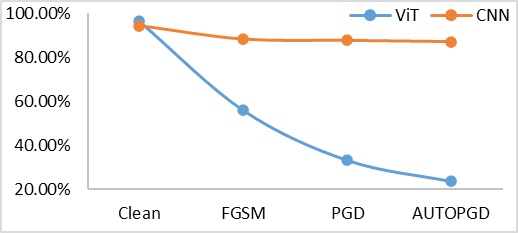}
\end{minipage}%
\begin{minipage}{.5\textwidth}
  \centering
  \includegraphics[width=7cm]{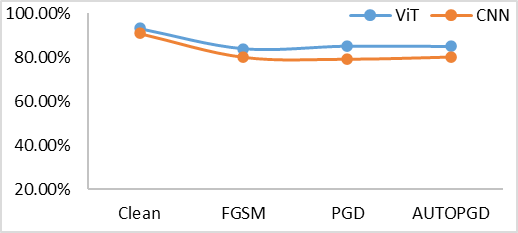}
\end{minipage}
\caption{Performance of distilled (blurred) ViT vs CNN (with original attack samples). Without defensive diffusion (left), and With defensive diffusion (right)}
\label{fig:teacher_student_Diff}
\end{figure}

We conducted a thorough comparison between the top-performing student CNN model and the teacher ViT model, specifically evaluating their performance on adversarial images and the output images obtained through defensive diffusion. The results, presented in the accompanying table, revealed a substantial decrease in accuracy for the ViT model when exposed to white box attacks. On the other hand, the CNN model consistently maintained a performance level comparable to that achieved with clean images. The accuracy of the ViT model witnessed a significant decline, plummeting from 88\% to approximately 20\% when confronted with white box attacks Figure \ref{fig:diff_cnn} and \ref{fig:teacher_student_Diff}. In contrast, the CNN model exhibited remarkable resilience, maintaining an accuracy rate of over 80\% even under adversarial attack. These outcomes highlight the robustness of the CNN model and its ability to withstand adversarial perturbations while preserving its high accuracy. The comparison results demonstrate the student CNN model's superior performance in the face of adversarial attacks, emphasizing its robustness and resilience compared to the ViT model. These findings contribute to our understanding of adversarial vulnerabilities across different model families and have implications for developing more robust defenses against adversarial attacks.

Furthermore, we proceeded to compare the accuracy of both models using the images generated from our method, specifically the blurred images. Surprisingly, the accuracy of the ViT model was effectively restored when evaluated on these blurred images. Notably, the CNN model exhibited robust performance, maintaining accuracy levels comparable to the ViT model. For instance, the ViT model's accuracy was restored from 2\% to over 80\% Figure \ref{fig:teacher_student_Diff} when subjected to the AutoPGD attack, which is known to be particularly challenging. This outcome serves as compelling evidence of the efficacy of our approach. The restoration of accuracy in the ViT model when evaluated on blurred images indicates the success of our defensive diffusion technique. Moreover, the fact that the CNN model's accuracy remains consistently high, even after undergoing the defensive diffusion process, suggests that the features learned by the CNN model are inherently more robust. This observation underlines the resilience of the CNN model in the face of adversarial perturbations and further validates the effectiveness of our proposed method.

\section{Conclusion and Future Work}
The paper's findings indicate that the diffusion model is successful in eliminating attack-agnostic adversarial noise from the image. Additionally, employing a lightweight student model proves to be effective in gray box attacks. The proposed defensive diffusion method outperforms the previous state-of-the-art approach (SEViT \cite{almalik2022self}) in most attack scenarios, achieving a significant improvement of over 2\% in accuracy compared to the PGD attack.
Our future work will involve:
\begin{enumerate}
\item Experimenting with different noise diffusion models, including denoising, masking, and snowing, in addition to the deblurring diffusion model.
\item Instead of fully reconstructing the original image through the reverse diffusion process, we will explore the use of slightly reconstructed images to obtain a slightly blurred image.
\item Employing an ensemble-based approach by training multiple diffusion models on subsets of training data, evaluating them on the original Vision Transformer (ViT), and aggregating the results using a majority voting technique.
\item In order to obtain more comprehensive and generalizable results, we will conduct tests on a diverse set of natural images.
\end{enumerate}

\bibliographystyle{unsrt} % We choose the "plain" reference style
\bibliography{references} % Entries are in the refs.bib file

\end{document}